\documentclass[10pt,twocolumn,letterpaper]{article}

\usepackage{cvpr}              % To produce the CAMERA-READY version
\usepackage{graphicx}
\usepackage{amsmath}
\usepackage{amssymb}
\usepackage{booktabs}
\graphicspath{ {./} }

\usepackage[pagebackref,breaklinks,colorlinks]{hyperref}

\usepackage[capitalize]{cleveref}
\crefname{section}{Sec.}{Secs.}
\Crefname{section}{Section}{Sections}
\Crefname{table}{Table}{Tables}
\crefname{table}{Tab.}{Tabs.}

\def\cvprPaperID{*****} % *** Enter the CVPR Paper ID here
\def\confName{CVPR}
\def\confYear{2022}

\begin{document}

%%%%%%%%% TITLE - PLEASE UPDATE
\title{Hands-Up: Leveraging Synthetic Data for Hands-On-Wheel Detection}

\author{Paul Yudkin\\
Datagen\\
{\tt\small paul.yudkin@datagen.tech}
% For a paper whose authors are all at the same institution,
% omit the following lines up until the closing ``}''.
% Additional authors and addresses can be added with ``\and'',
% just like the second author.
% To save space, use either the email address or home page, not both
\and
Eli Friedman\\
Datagen\\
{\tt\small eli.friedman@datagen.tech}
\and
Orly Zvitia\\
Datagen\\
{\tt\small orly.zvitia@datagen.tech}
\and
Gil Elbaz\\
Datagen\\
{\tt\small gil@datagen.tech}
}
\maketitle

%%%%%%%%% ABSTRACT
\begin{abstract}
    Over the past few years there has been major progress in the field of synthetic data generation using simulation based techniques. These methods use high-end graphics engines and physics-based ray-tracing rendering in order to represent the world in 3D and create highly realistic images. Datagen has specialized in the generation of high-quality 3D humans, realistic 3D environments and generation of realistic human motion. This technology has been developed into a data generation platform which we used for these experiments. 
    This work demonstrates the use of synthetic photo-realistic in-cabin data to train a Driver Monitoring System that uses a lightweight neural network to detect whether the driver’s hands are on the wheel. We demonstrate that when only a small amount of real data is available, synthetic data can be a simple way to boost performance. Moreover, we adopt the data-centric approach and show how performing error analysis and generating the missing edge-cases in our platform boosts performance. This showcases the ability of human-centric synthetic data to generalize well to the real world, and help train algorithms in computer vision settings where data from the target domain is scarce or hard to collect.

\end{abstract}

%%%%%%%%% BODY TEXT
\section{Introduction}
\label{sec:intro}
Currently, most vehicles on the road are driven by humans. People are prone to distractions while driving, which is the cause of 15\% of the injury-causing accidents in the US \cite{usdot}. In the next few years the European regulations will require car manufacturers to gradually include new safety technologies, such as Driver Monitoring Systems (DMS) in vehicles \cite{eu_dms}.  
% These regulations will spread globally to China and the USA over time.
In addition, the European NCAP has started requiring driver monitoring features in order to qualify for a 5-star safety rating \cite{ncap}, raising the urgency of development of driver monitoring systems.

In this paper, we develop a driver monitoring system to detect when a driver’s hands leave the wheel. Such a system could be useful in multiple DMS tasks, such as to raise an alert when the driver is distracted.
As with many deep learning projects, the challenge here is data. It is difficult to collect many images of drivers in a vehicle, and while some datasets do exist, they are limited in the number of drivers, vehicle types, behaviors, and camera models that they use. In addition, tagging tens of hours of videos in a consistent way is challenging. Synthetic data provides an alternative solution that, once developed, can be used to create significant variance with little manual effort. 

We developed a synthetic data platform that renders highly realistic scenes of drivers in cars. Our synthetic data platform allows varying the camera position and type, scene lighting, and driver behavior (e.g., falling asleep, looking around, drinking, texting etc.,). It includes pixel perfect ground truth and 3D annotations so that no manual tagging is required. 
Our two main contributions in this work are: 1) We demonstrate how using synthetic data along with a very small amount of real examples can boost performance relative to using the same amount of only-real data. 2) We show in practice a complete iteration of a data-centric approach using our platform to generate a specific edge case that we were lacking in the training dataset.

\section{Literature Review}
Increasingly, deep learning systems are being used to implement DMS systems. Kose et al.\ \cite{real_time_dms} use a convolutional network to classify distracted driving, and Rangesh et al.\ \cite{handynet} build a predictor to segment and localize the driver's hands. The most important part of any AI based DMS system is collecting data. The Drive \& Act dataset \cite{drive_and_act} contains 15 subjects, but the annotated behaviors focus on tasks a driver might perform in an autonomous vehicle instead of things a driver would do while driving.
% The DriPE dataset \cite{human_pose} contains 19,000 images of drivers with manually annotated 2D body keypoints.
The DMD dataset \cite{dmd_dataset} is a comprehensive dataset, containing 37 drivers and 42 hours worth of video data. It contains videos of both real driving scenarios as well as driving in simulators, and includes annotated driving behaviors. It is, to our knowledge, the most comprehensive public dataset for research on DMS systems, which is why we chose it as our comparison dataset. 
Synthetic data is becoming more prelavent as the need for data grows. Tremblay et al.\ \cite{synth_nvidia} train an object detection network using synthetic data and then finetune on real data. Sengupta et al.\ \cite{smpl_synth} use a synthetic model to help regress the pose of a human, which could be used as part of a pipeline to first detect the pose of the driver's hands and then whether the hands are on the wheel. 
However we opt for a single stage pipeline that directly predicts whether the hands are on the wheel.
\section{Method}
\subsection{Real Data Preparation}
We chose the DMD dataset \cite{dmd_dataset} as our real  dataset because it contains a large number of drivers, driving scenarios, camera angles, and has a wide variety of tagged behaviors, including whether the hands are on the wheel. 
% The dataset is divided into four groups of drivers, each group containing five drivers, and each driver participating in four recorded driving sessions–three in an actual car and one in a simulator.
We split the dataset into a train, validation, and test sets based on the identity of the drivers in the dataset. In total, the dataset contains 651k frames, of which we use 531k for training, 47k for validation, and the rest for test. 
The drivers are recorded using three cameras--one facing the driver's head, one facing the driver's body, and one facing the driver's hands. We use the camera facing the driver's body because a side view of the wheel offers a clearer perspective whether the hands are on the wheel.
The dataset is not balanced between the hands. Drivers in countries that drive on the left side of the road will typically perform other actions with their right hand while the left hand remains on the wheel. This bias can be seen in Table \ref{table:datasets_breakdown}.
\begin{table}[h!]
\centering
\begin{tabular}{|| c  c |c |c||} 
\hline
Left & Right Hand & Synthetic & Real\\[0.5ex] 
 \hline\hline
 On & On & 5642 (50\%) & 214192 (32.8\%)\\
 On & Off & 3546 (31.4\%) & 304102 (46.7\%)\\
 Off & On & 2014 (17.8\%) & 122416 (18.8\%)\\
 Off & Off & 82 (0.7\%) & 10579 (1.6\%)\\
 \hline
 & Total & 11,284 & 651,289\\
   \hline\hline
\end{tabular}
\caption{ Label distribution in real and synthetic datasets.}
\label{table:datasets_breakdown}
\end{table}

% \begin{figure}[h]
%     \centering
%     \includegraphics[width=0.5\textwidth]{imgs/dataset_breakdown}
%     \caption{Tag breakdown for the real (left) and synthetic (right) datasets}
%     \label{fig:datasets_breakdown}
% \end{figure}

\subsection{Synthetic Data Preparation}
\begin{figure}[ht]
\centering
\includegraphics[width=0.35 \textwidth]{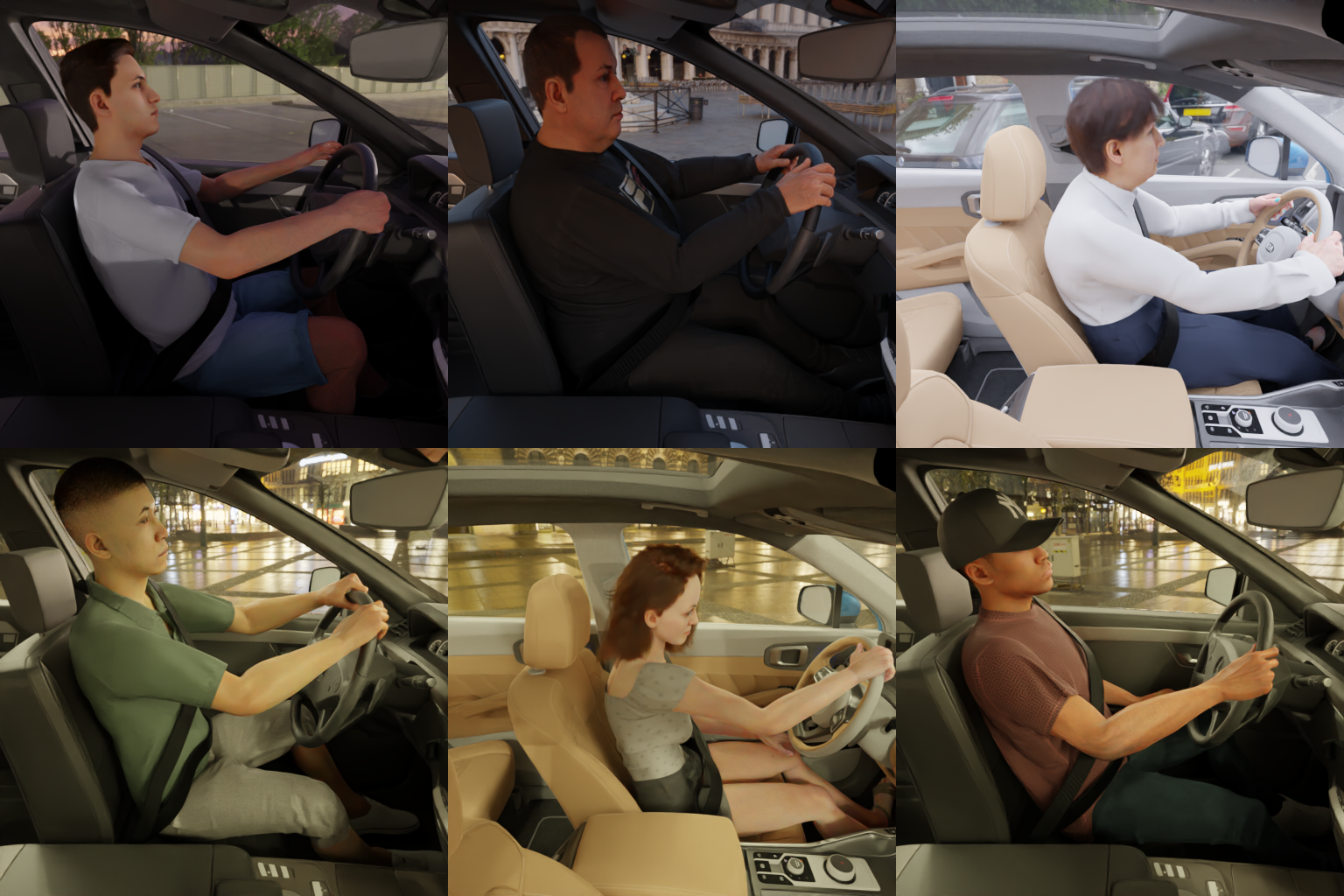}
\caption{Sample images from our synthetic dataset}
\label{fig:synthetic_samples}
\end{figure}

We use the Datagen synthetic data platform to generate a diverse video dataset composed of different drivers who perform various actions in different vehicles. Among multiple camera views available, we render the scene using a camera focused on the driver's body, a similar viewpoint as the real data. Each scene is 10 seconds long and is rendered at 15 frames per second. Each image resolution is 256x256 and includes hand and body keypoints, and wheel keypoints.  See Figure \ref{fig:synthetic_samples} for some RGB examples from the synthetic dataset.

To maximize variance in our dataset we generated diverse sequences with respect to the following aspects:
1) Environment -  Our dataset includes various car types including large and medium SUVs and sedan type vehicles. The interior areas in the car differ to allow variance including seat types, wall colors and, especially important for our task, different wheel types. 2) Demographics - We used ten different drivers with different ethnicity, age and genders. 3) Behaviors - We generate multiple behaviors such as falling asleep, turning around, texting, one handed driving, and regular two handed driving.  4) Scene - We generate all sequences with a random background and lighting condition--daylight, evening light, or night. In total we generate 146 sequences.

For each frame we separately label each hand as being on or off the steering wheel. The availability of 3D key points from our platform makes the hands-on-wheel labeling almost a trivial task. We simply calculate the distance from the wheel to the closest point on each hand and consider the hand to be on the wheel if it is closer than 3 cm.
\subsection{Synthetic Data Splits}
In order to balance the distribution of labels in our dataset, we undersampled the synthetic dataset and removed labels with both hands on wheel. In total, the synthetic dataset contains 11,284 unique images. We split our train, validation, and test sets based on the driver identity. Our training set contains 8,834 images. The validation set consists of 2,450 images following the same proportions as the train split.

\begin{figure}
    \centering
    \includegraphics[width=0.45 \textwidth]{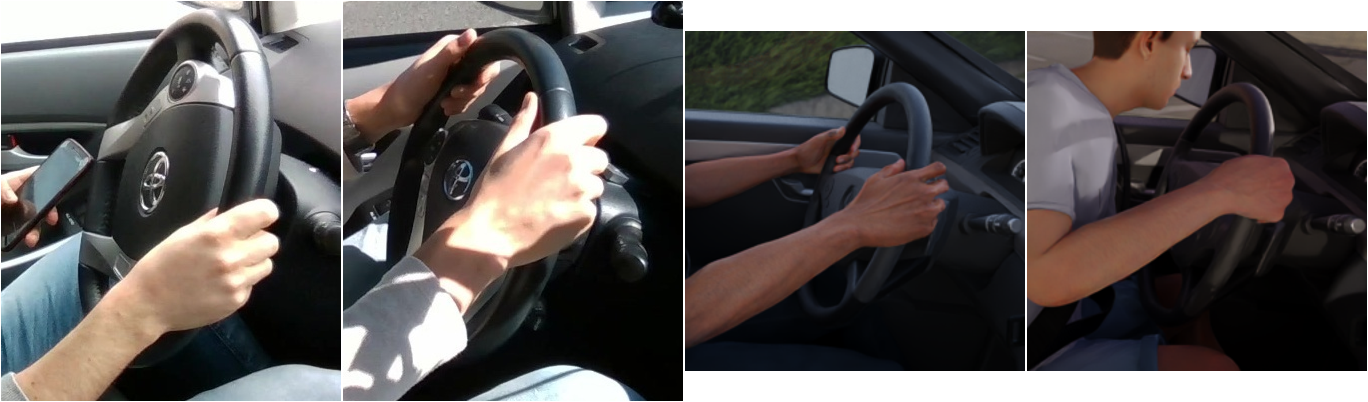}
    \caption{Examples from the real and synthetic datasets after cropping around the wheel}
    \label{fig:samples_cropped}
\end{figure}

\subsection{Pre-processing}
We wanted to eliminate background distractions from the model. Therefore, we manually crop both the real images and synthetic images around the wheel, so that only the wheel and hands are visible without any extraneous details. See Figure \ref{fig:samples_cropped} for some examples from the real and synthetic datasets.

\subsection{Model Architecture}
We choose the lightweight MobileNetV3 \cite{mobile_net} architecture as backbone for all our experiments considering the real-time nature of our task. We replaced the classification head with two binary classification heads each containing two fully connected layers activated with ReLU \cite{relu} and a final fully connected layer with a sigmoid activation. The two classification heads predict, respectively, whether the left or right hand is on the wheel.

\section{Experiments}
We conduct two types of experiments to demonstrate the added value of easily reachable synthetic data. 
1) We compare a model trained solely on the DMD real data with multiple models trained on synthetic data and fine-tuned on varying amounts of real data mixed with synthetic data. We assume tagging a small amount of real data is feasible in most cases and preferable over tagging hundreds of hours.
2) We show that one can boost performance by applying a data-centric iteration--searching the test errors for edge cases that are missing from the dataset and adding them.

We evaluate our performance with AUC scores for each of the hands.

\subsection{Training \& Fine-tuning}

We refer to a model trained on DMD as our reference, and compare it to a model trained mainly on synthetic data and boosted with very limited amount of real data. We consider two methods for using the synthetic data. 1) Train on the synthetic data alone and 2) Train on the synthetic dataset followed by fine-tuning with a mix of real and synthetic data. We train all models using Adam optimizer\cite{Adam} with $\beta_1  = 0.9$,  $\beta_2  = 0.9999$, batch size of 128, weight decay of 0.1, and initial learning rate of 0.0001.

\begin{table}[h!]
\centering
\begin{tabular}{||c | c|c |c||} 
\hline
Experiment & \small{\# Train Images} & Left \footnotesize{AUC} & Right \footnotesize{AUC}\\[0.5ex] 
 \hline\hline
   Real Only & 531k & 0.9813 & 0.9941\\
   Synth Only & 8,800 & 0.7226 & 0.7581\\
   Synth+100 Real & 8,900 & 0.8769 & 0.9045\\
   Synth+200 Real & 9,000 & 0.9139 & 0.9389\\
   Synth+300 Real & 9,100 & 0.9251 & 0.9475\\
   Synth+400 Real & 9,200 & 0.9369 & 0.9530\\
   \hline\hline
\end{tabular}
\caption{AUC scores for models trained on synthetic dataset, real dataset, and tested on the real dataset}
\label{table:pure_synth_real}
\end{table}

\begin{figure}[t]
    \centering
    \begin{subfigure}[t]{0.2 \textwidth}
        \centering
        \includegraphics[width=\textwidth]{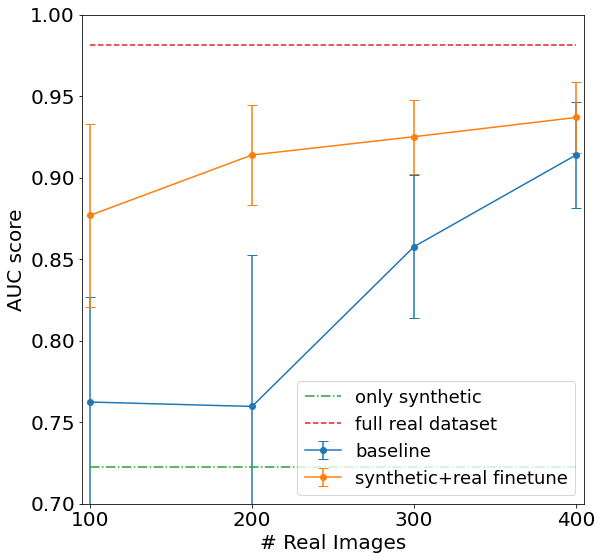}
        \caption{Left Hand}
        \label{fig:finetune_left}
    \end{subfigure}%
    \begin{subfigure}[t]{0.2 \textwidth}
        \centering
        \includegraphics[width=\textwidth]{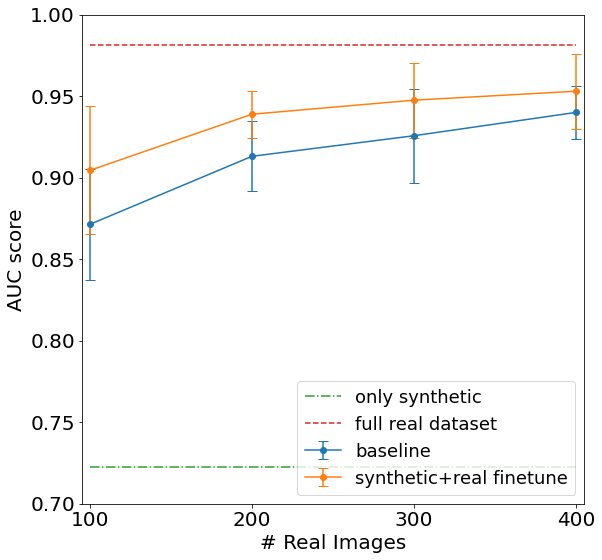}
        \caption{Right Hand}
        \label{fig:finetune_right}
    \end{subfigure}
    \caption{AUC comparison between baseline results (blue) and synthetic fine-tune results (orange) for the left hand (\ref{fig:finetune_left}) and right hand (\ref{fig:finetune_right})}
    \label{fig:finetune_results}
\end{figure}

We would expect that due to the difference in dataset sizes, as well as because of the domain gap, it is to be expected that a model trained on synthetic dataset alone performs worse than one trained on the full dataset. This is exactly what we see in Table \ref{table:pure_synth_real}, the real dataset performs significantly better than the model trained on the synthetic dataset.
We also sub-sample the DMD dataset to create four small datasets with only 100, 200, 300, and 400 frames in them, respectively. We create the datasets by choosing five drivers and sample 5, 10, 15, and 20 frames from each of the four videos that each driver appears in. We train the model on synthetic data and then fine-tune it for 2,500 batches using a mix of synthetic data and real data from the small datasets. When fine-tuning, we make sure that each batch contains an even mix of synthetic data and real data. Without this technique, the network forgets its initial training on the synthetic data. 
We demonstrate the effectiveness of synthetic data by comparing to models trained solely on each of these small real datasets.
We ran the experiment 11 times on different sub-sampling splits of the data and show the AUC mean and standard deviation on the real test set for each small dataset in Figure \ref{fig:finetune_results}. The large error intervals are caused by the sensitivity to the small amount of data in each split.

Since the right hand is clearly visible in most frames, the network receives enough variations to learn. Therefore, the synthetic data does not improve the model as much over the real data. However, for the left hand, which is often occluded by the right hand, the lack of variations is noticeable, and the synthetic data provides a clear improvement. This is especially true when there are fewer than 200 images in the real training set and the results improve from 0.76 AUC to 0.91. In this case, the network did not have the opportunity to see all the different edge cases that are only present in the synthetic data.

\subsection{Data-Centric Iteration}

\begin{figure}[t]
    \centering
    \begin{subfigure}[t]{0.125\textwidth}
        \centering
        \includegraphics[width=0.95\textwidth]{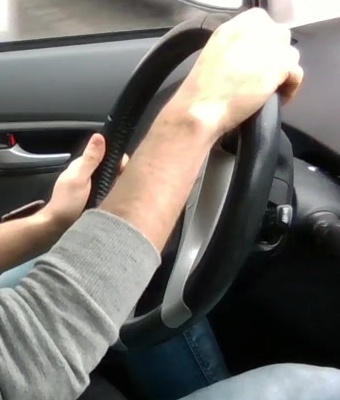}
        \caption{Occlusion}
        \label{fig:occluded_hands}
    \end{subfigure}%
    \begin{subfigure}[t]{0.125\textwidth}
        \centering
        \includegraphics[width=0.95\textwidth]{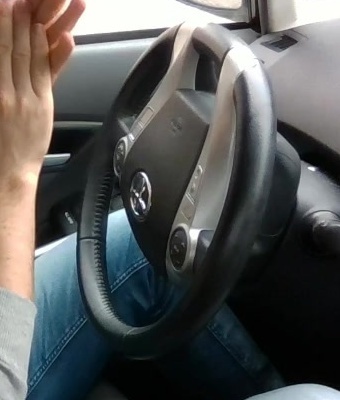}
        \caption{Both hands off}
        \label{fig:both_hands_off}
    \end{subfigure}%
    \begin{subfigure}[t]{0.125\textwidth}
        \centering
        \includegraphics[width=0.95\textwidth]{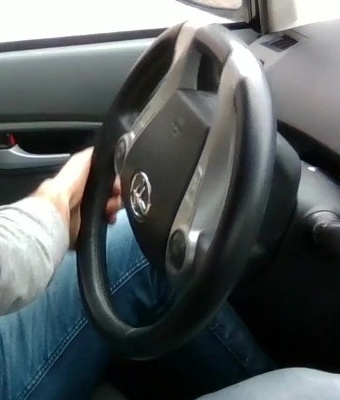}
        \caption{Hand on opposite side}
        \label{fig:hand_wrong_side}
    \end{subfigure}%
    \begin{subfigure}[t]{0.125\textwidth}
        \centering
        \includegraphics[width=0.95\textwidth]{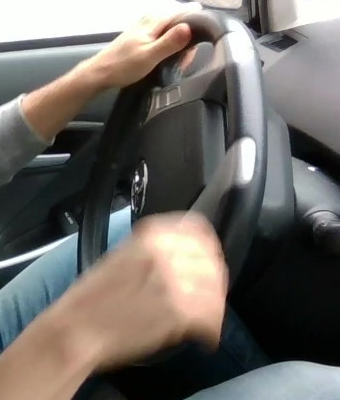}
        \caption{Blur}
        \label{fig:blur}
    \end{subfigure}
    \caption{Some examples of common types of errors. (\ref{fig:occluded_hands}) (\ref{fig:both_hands_off}) Left hand classified as on (\ref{fig:hand_wrong_side}) Left hand classified as on, Right hand classified as off (\ref{fig:blur}) Right hand classified as on}
    \label{fig:error_cases}
\end{figure}

\begin{figure}[t]
    \centering
    \includegraphics[width=0.45\textwidth]{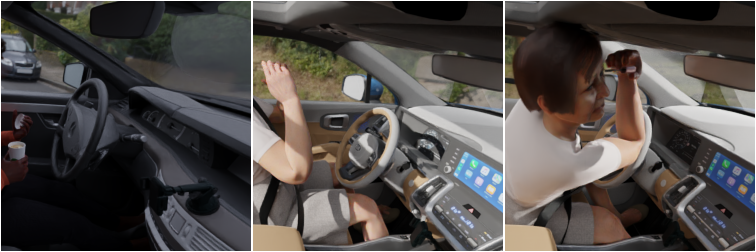}
    \caption{Some new frames with both hands off that were added to the synthetic dataset after the data-centric iteration.}
    \label{fig:new_frames}
\end{figure}

In addition to iterating on the model, we also iterate on the dataset. We use our base model, which was trained only on synthetic data (results in second row of Table \ref{table:pure_synth_real} ), and we visually analyze the errors in the DMD validation split.
The majority of misclassifications can be divided into several specific categories:
1) Occlusion - Hands overlap in the image, so the network has a hard time telling whether the left hand is on or off the wheel (See Figure \ref{fig:occluded_hands})
2) Both Off - This is an uncommon case, so the network has a harder time classifying this case correctly (See Figure \ref{fig:both_hands_off})
3) Opposite Side - The right hand is on the left side of the wheel or vice versa. The network will classify the left hand as "on" and the right hand as "off" (See Figure \ref{fig:hand_wrong_side})
4) Blur - The video is blurry when the hand is in motion and so it's unclear if the hand on wheel or not (See Figure \ref{fig:blur})
Based on our failure analysis we generated sequences with both hands off wheel (total of 450 images). Examples for the generated frames is shown in Figure \ref{fig:new_frames}. After retraining, performance on this specific scenario increased, with the recall and precision jumping from 0.77 and 0.98 to 0.85 and 0.99, respectively.

%\begin{figure}[h]
%    \centering
%    \begin{subfigure}[h]{0.5\textwidth}
%        \centering
%        \includegraphics[width=0.8\textwidth]{imgs/errors/left_%hand_errors.png}
%        \caption{Left hand errors}
%        \label{fig:left_hand_errors}
%    \end{subfigure}
%    \begin{subfigure}[h]{0.3\textwidth}
%        \centering
%        \includegraphics[width=0.8\textwidth]{imgs/errors/right%_hand_errors.png}
%        \caption{Right hand errors}
%        \label{fig:right_hand_errors}
%    \end{subfigure}
%    \caption{Most common classification errors for the left %(\ref{fig:left_hand_errors}) and right (\ref{fig:right_hand_errors}) hands}
%    \label{fig:common_errors}
%\end{figure}

% \begin{figure}[h]
%     \centering
%     \includegraphics[width=0.3\textwidth]{imgs/errors/both_hand_errors.png}
%     \caption{Most common classification errors. Fist row - Left On predicted off, left
%     , Second row - Right hand.}
%     \label{fig:both_hand_errors}
% \end{figure}

\section{Conclusion}
In this paper, we simulate a situation in which only a small amount of real data is available and demonstrate how synthetic data can compensate for the missing real data. We show that mixing real and synthetic data outperforms training on a small amount of real data alone. Introducing small amount or real data to a synthetic-first model boosts performance by compensating the domain gap reaching almost the same results as training on a large amount of real data. Furthermore, we followed the data-centric approach applying a single data improvement iteration leveraging our configurable platform that successfully improved our model emphasizing the great potential or incorporating synthetic data in real-life models.
\section{Future Work}
Further research is required to improve results when training solely with synthetic data. We believe that adopting more iterations of the data centric approach will improve results. This involves iterations of failure analysis and updating the training dataset appropriately. Supplementing our model with additional information, such as depth maps or sequential frame information could also improve results. Another interesting experiment would be to compare results among different cameras and explore combinations of multiple cameras to help compensate for occluded areas. This could also support a pipeline that involves identifying the location of the hands individually rather than classifying the image directly. Additionally, we plan to utilize our 3D pixel perfect key-points to solve hands-on-wheel problem using pose-estimation of the hands. A final interesting direction involves the use of unsupervised pretraining to ensure that the feature distributions of the synthetic and real data are similar, thus overcoming the domain gap. 

%%%%%%%%% REFERENCES
{\small
\bibliographystyle{ieee_fullname}
\bibliography{egbib}
}

\end{document}